\documentclass[letterpaper, 10 pt, conference]{ieeeconf}  

\IEEEoverridecommandlockouts                              
\usepackage[table,xcdraw,dvipsnames]{xcolor}
\usepackage{subcaption}
\usepackage{graphicx}
\usepackage{textcomp}
\usepackage{balance}
\usepackage{amssymb}
\usepackage{amsmath}
\usepackage{pifont}

\newcommand{\xmark}{\text{\ding{55}}}

\usepackage{hyperref} 
 \hypersetup{ 
     colorlinks=true, 
     linkcolor=black, 
     filecolor=blue, 
     citecolor = black,       
     urlcolor=blue, 
     } 
\title{\LARGE \bf
QueensCAMP: an RGB-D dataset for robust Visual SLAM
}

\author{Hudson M. S. Bruno$^{1}$, Esther L. Colombini$^{1}$ and Sidney N. Givigi Jr.$^{2}$
\thanks{$^{1}$Institute of Computing, Universidade de Campinas
Campinas, SP, Brazil. {\tt\small hudson.bruno@ic.unicamp.br}}%
\thanks{$^{2}$School of Computing, Queen’s University. Kingston, ON, Canada.{\tt\small sidney.givigi@queensu.ca}}
}

\begin{document}

\maketitle
\thispagestyle{empty}
\pagestyle{empty}

\begin{abstract}

Visual Simultaneous Localization and Mapping (VSLAM) is a fundamental technology for robotics applications. While VSLAM research has achieved significant advancements, its robustness under challenging situations, such as poor lighting, dynamic environments, motion blur, and sensor failures, remains a challenging issue. To address these challenges, we introduce a novel RGB-D dataset designed for evaluating the robustness of VSLAM systems. The dataset comprises real-world indoor scenes with dynamic objects, motion blur, and varying illumination, as well as emulated camera failures, including lens dirt, condensation, underexposure, and overexposure. Additionally, we offer open-source scripts for injecting camera failures into any images, enabling further customization by the research community. Our experiments demonstrate that ORB-SLAM2, a traditional VSLAM algorithm, and TartanVO, a Deep Learning-based VO algorithm, can experience performance degradation under these challenging conditions. Therefore, this dataset and the camera failure open-source tools provide a valuable resource for developing more robust VSLAM systems capable of handling real-world challenges.

\end{abstract}

\section{INTRODUCTION}

The process of Visual Odometry (VO) involves estimating an agent's position and orientation (pose) based solely on visual information obtained from cameras attached to it~\cite{tutorial-vo}. VO is a crucial component of Visual Simultaneous Localization and Mapping (VSLAM). VSLAM has become a fundamental technology in robotics, enabling systems to navigate and understand their environment using visual data autonomously. It is widely used in robotics due to its low cost and lightweight nature \cite{uav-application}.

However, the deployment of these systems in real-world scenarios presents multiple challenges, especially because VO and VSLAM algorithms are prone to fail under challenging situations, such as textureless areas, inadequate illumination, motion blur, varying weather conditions, and camera lens failures \cite{overview-dl-vslam}. Consequently, ensuring the robustness of these systems is considered one of the most challenging tasks in the field today \cite{robust-slam-systems}.

The evaluation and improvement of VSLAM algorithms depend on the availability of datasets that simulate real-world challenges. While widely employed benchmarks such as KITTI \cite{kitti-dataset}, TUM RGB-D \cite{tum-rgbd}, and Euroc MAV \cite{euroc-mav} offer diverse environments and motion patterns, these datasets lack the complexity needed to fully assess VSLAM performance under challenging scenarios.

\begin{figure}[t]
\centering
\subfloat[][Original Image.]{
\includegraphics[width=0.2\textwidth]{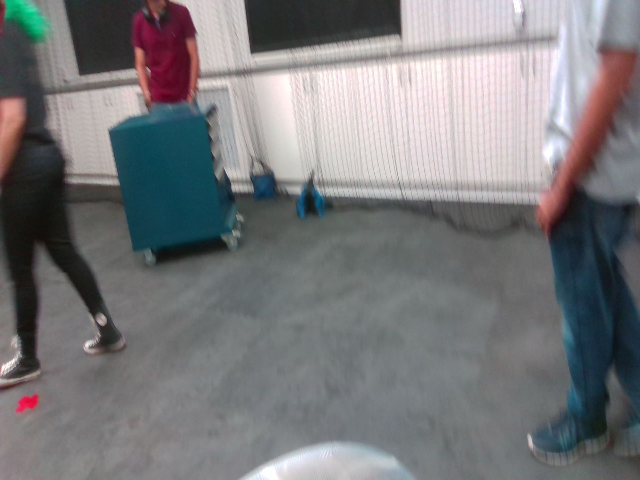}
\label{fig:original}}
\qquad
\subfloat[][Depth Image.]{
\includegraphics[width=0.207\textwidth]{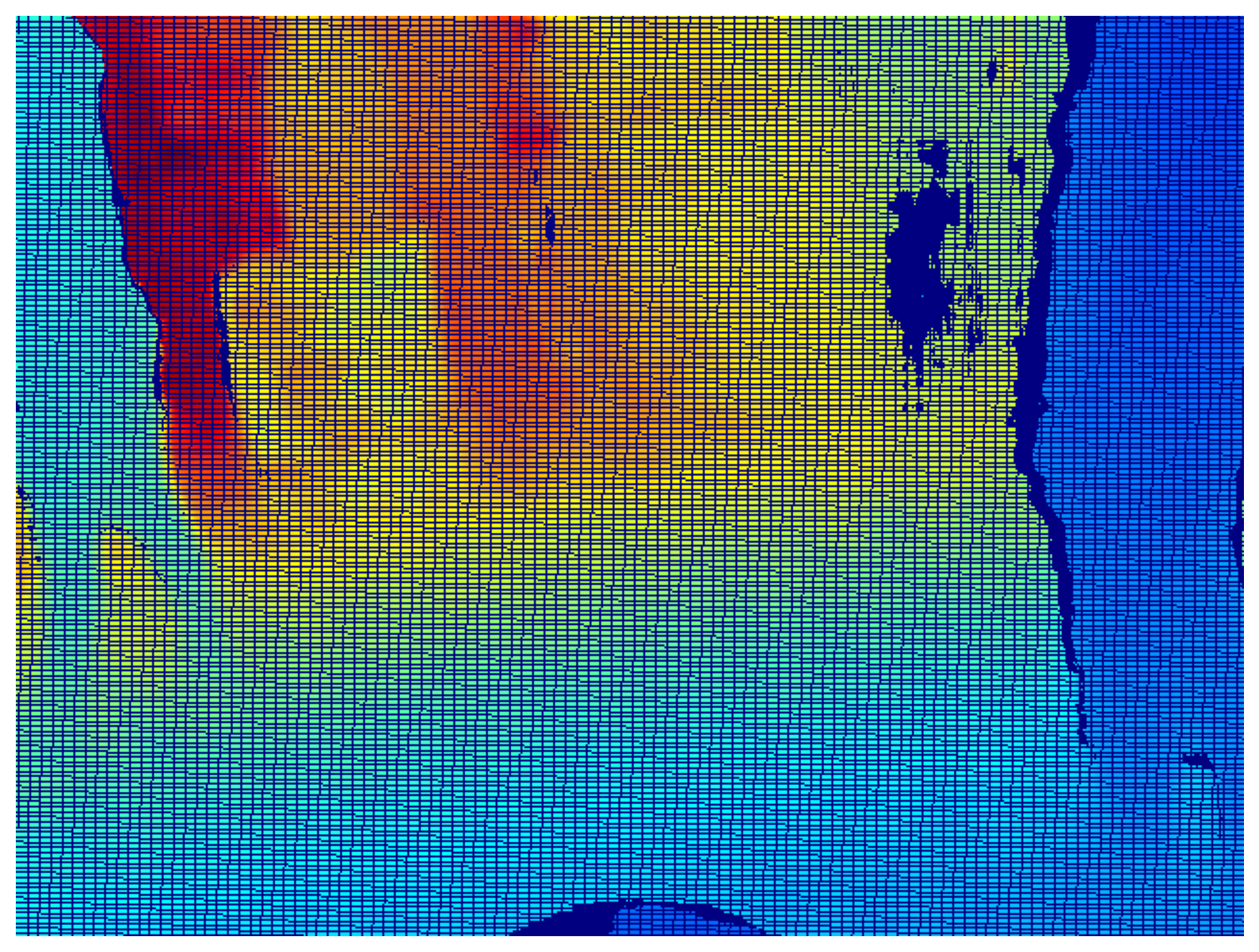}
\label{fig:sample-depth}}

\subfloat[][Camera Underexposure.]{
\includegraphics[width=0.2\textwidth]{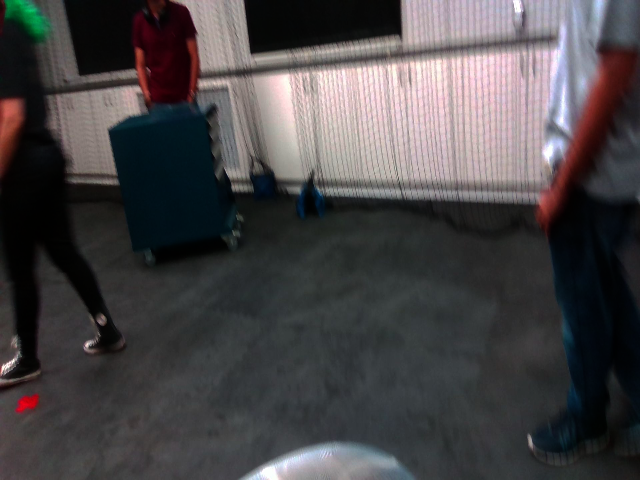}
\label{fig:underexposure}}
\qquad
\subfloat[][Camera Overexposure.]{
\includegraphics[width=0.2\textwidth]{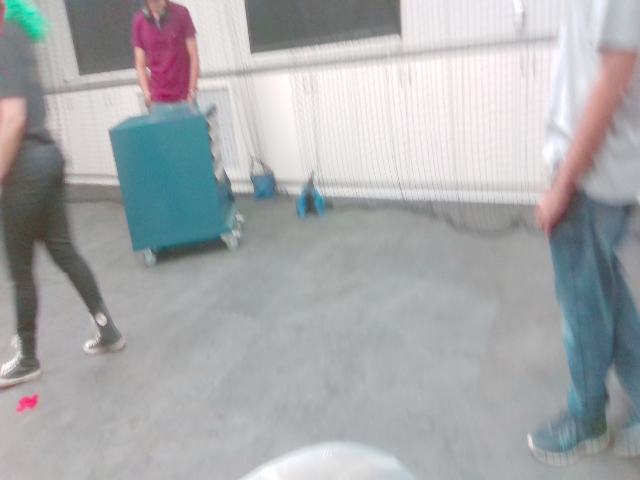}
\label{fig:overexposure}}

\subfloat[][Lens Breakage.]{
\includegraphics[width=0.2\textwidth]{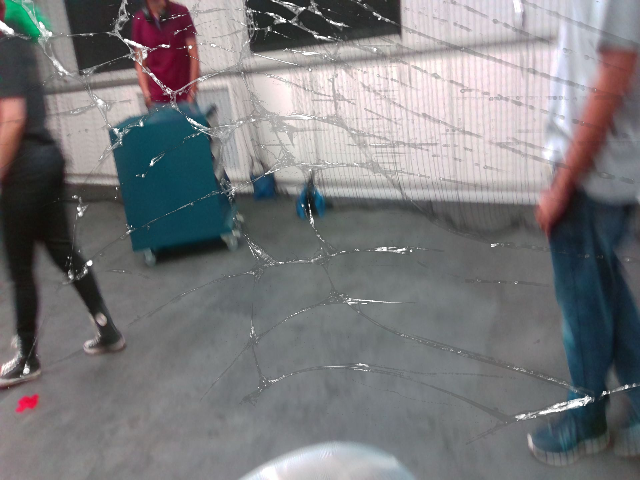}
\label{fig:breakage}}
\qquad
\subfloat[][Wet Lens.]{
\includegraphics[width=0.2\textwidth]{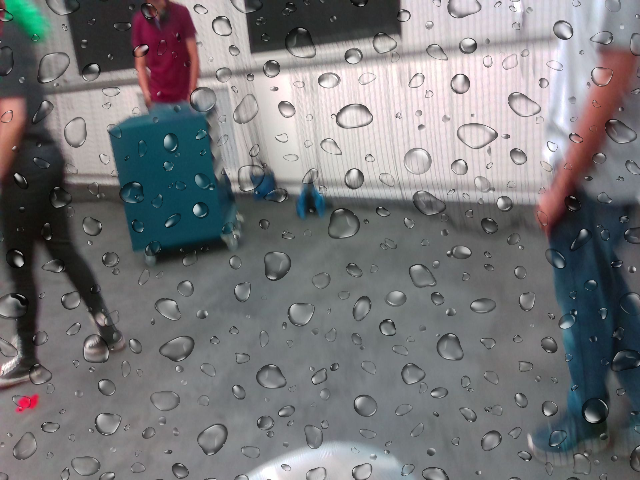}
\label{fig:wet}}

\subfloat[][Lens Condensation.]{
\includegraphics[width=0.2\textwidth]{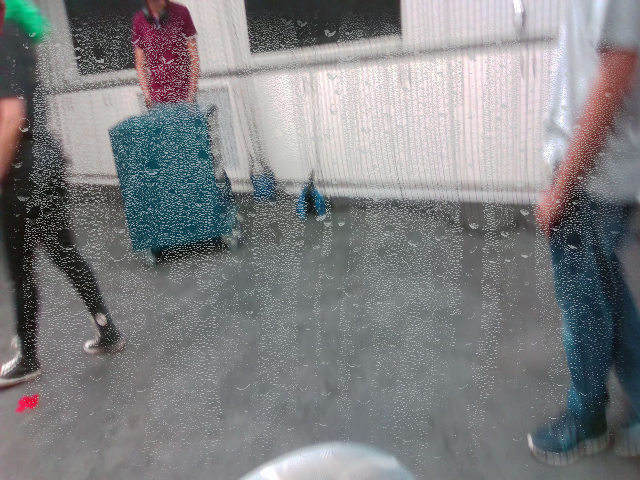}
\label{fig:condensation}}
\qquad
\subfloat[][Lens Dirt.]{
\includegraphics[width=0.2\textwidth]{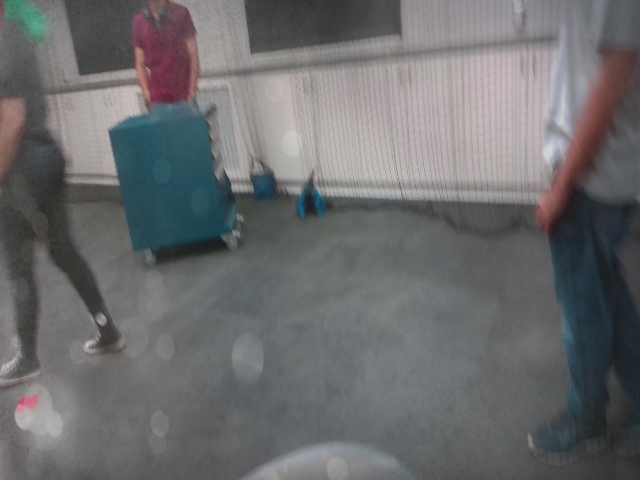}
\label{fig:dirt}}

\caption{A sample of the captured data and the emulated failures used to create adverse conditions in the dataset.}
\label{fig:failures}
\end{figure}

To address these gaps, this paper introduces an RGB-D VSLAM dataset that includes not only easy-to-capture challenges such as dynamic objects, varied lighting conditions, and motion blur but also incorporates new scenarios involving lens failures. By simulating issues like broken lenses, wet lenses, condensation, dirty lenses, and exposure problems, this dataset provides more challenging situations for the robustness evaluation of VSLAM systems. Fig. \ref{fig:failures} shows a sample of the dataset and all types of lens failures included in the dataset.


Therefore, the main contributions of this work are:

\begin{enumerate} 
\item A comprehensive dataset of real-world indoor RGB-D scenes, featuring dynamic objects, motion blur, and varying illumination, designed for the robust evaluation of VSLAM algorithms. The dataset is publicly available\footnote{\url{https://larocs.github.io/queenscamp-dataset}};
\item The inclusion of emulated camera failures consistently applied across the dataset, enabling a thorough evaluation of VSLAM robustness under challenging lens failure conditions; 
\item Open-source scripts for raw data post-processing and camera failure emulation, made freely available at github\footnote{\url{https://github.com/larocs/queenscamp-dataset}\label{footnote:repo}}, allowing for easy integration and extension by the research community. 
\end{enumerate}

The remainder of this paper is organized as follows: Section~\ref{sec:related-work} presents related works that propose datasets for VSLAM evaluation, Section~\ref{sec:methodology} gives details about the design of the dataset and how we evaluated the collected data using two open-source VO and VSLAM algorithms. Section~\ref{sec:results} contains the results obtained with the experiments using the two algorithms and our conclusions are in Section~\ref{sec:conclusion}.

\section{RELATED WORK}
\label{sec:related-work}
The robustness of VSLAM systems under challenging conditions has been a longstanding concern for researchers in robotics~\cite{robust-slam-systems}. To evaluate VSLAM systems, multiple datasets have been introduced. The most commonly used datasets for evaluating these systems are KITTI~\cite{kitti-dataset}, TUM RGB-D~\cite{tum-rgbd}, and Euroc MAV~\cite{euroc-mav}. 

The KITTI dataset~\cite{kitti-dataset} consists of $22$ stereo image sequences covering a total distance of $39.2$ km, captured from a moving car. However, KITTI is limited in its motion patterns, as the vehicle predominantly moves forward and does not exhibit significant upward or backward motion. In contrast, the TUM RGB-D dataset~\cite{tum-rgbd} offers a benchmark for evaluating RGB-D SLAM systems, with $39$ sequences recorded in two indoor environments using an RGB-D camera mounted on a Pioneer 3 robot and a handheld camera. Moreover, the Euroc MAV dataset~ \cite{euroc-mav} is designed for the assessment of visual-inertial SLAM and 3D reconstruction, providing $11$ sequences that range from slow flights under favorable visual conditions to dynamic flights with motion blur and poor illumination. This dataset focuses on small-scale indoor scenes with six degrees of freedom (DoF).

Although the KITTI, Euroc MAV, and TUM RGB-D datasets cover a variety of scenes and camera motions, they do not address some of the more challenging scenarios, such as adverse weather conditions, featureless areas, and poor illumination. To tackle these limitations, other specialized datasets have been proposed. For instance, a benchmark for outdoor visual localization evaluation, under changing illumination and weather, was proposed in~\cite{benchmark-outdoor-changing-conditions}. Similarly, Carlevaris-Bianco et al.~\cite{north-campus} proposed a dataset containing changing environments (indoor and outdoor), moving obstacles, changing lighting, varying viewpoints, and seasonal and weather changes. Moreover, in~\cite{oxford-robotcar} a dataset with more than $20$ million images was provided with varying weather conditions, including heavy rain, night, direct sunlight, and snow.

Among the most challenging datasets is TartanAir \cite{tartanair}. This dataset is collected in photo-realistic simulation environments with various lighting conditions, weather, dynamic objects, and randomized camera motion. TartanAir includes $30$ environments with $1,037$ long motion sequences, covering both indoor and outdoor settings. The dataset leverages the Unreal Engine, with data collected using Microsoft's AirSim plugin \cite{airsim}.

Additionally, the OpenLORIS-Scene dataset \cite{openloris} was proposed for lifelong SLAM applications in indoor environments. They have used RGB-D cameras and inertial sensors to capture data in $5$ different scenes. The dataset contains dynamic movement, sensor degradation, changed viewpoints, objects, and illumination.

While these datasets provide valuable resources for evaluating VSLAM algorithms and present challenging conditions such as motion blur, adverse weather, and dynamic objects, they do not account for lens failures, such as breakage or condensation. To the best of our knowledge, this is the first VSLAM dataset to incorporate emulated failures alongside challenging conditions to evaluate the robustness of the systems.

\section{METHODOLOGY}
\label{sec:methodology}
In this section, we explain how the dataset was built, including the materials used to collect the data and the scripts applied to inject failures. We also present the algorithms used to evaluate the collected data and the metrics used to assess the estimated trajectories.

\subsection{Data collection}

\begin{figure}
    \centering
    \includegraphics[width=0.8\linewidth]{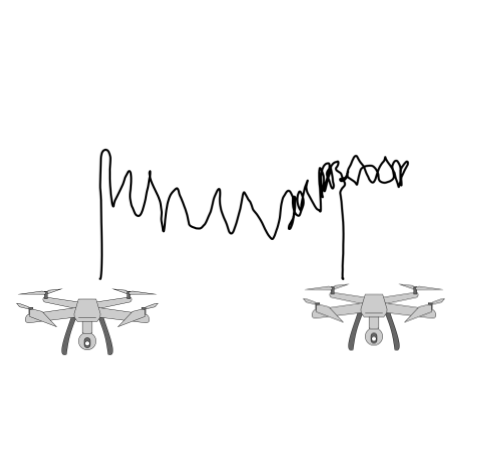}
    \caption{Sample of a six DoF ground-truth trajectory collected with the Vicon motion capture system.}
    \label{fig:gt-sample}
\end{figure}

\begin{figure}
    \centering
    \includegraphics[width=0.8\linewidth]{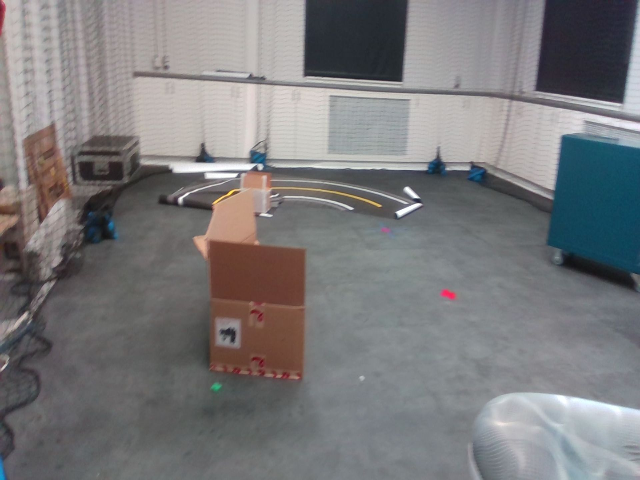}
    \caption{The environment used to collect the data.}
    \label{fig:environment}
\end{figure}

\begin{figure}
    \centering
    \includegraphics[width=\linewidth]{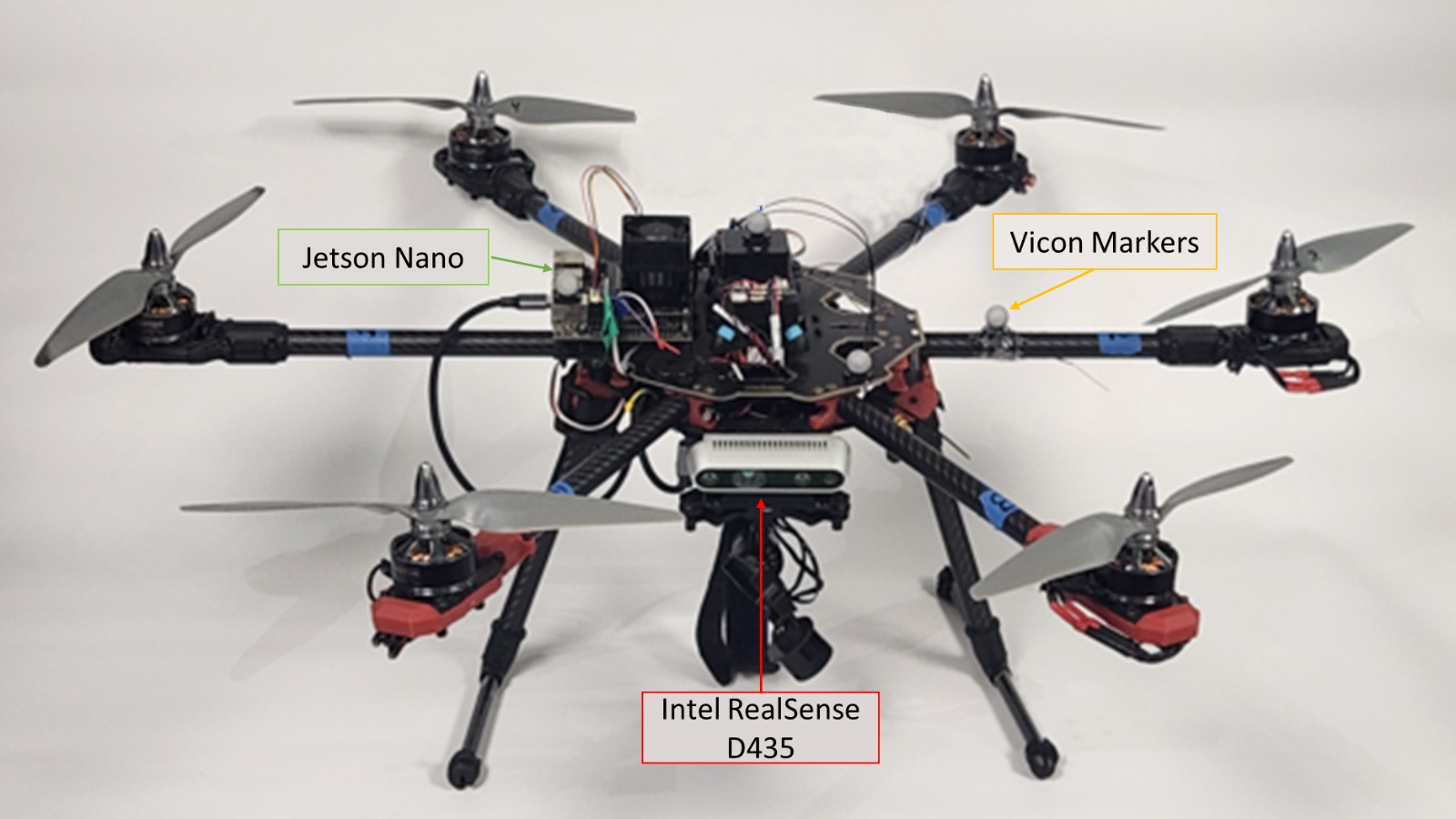}
    \caption{The vehicle used to capture the data.}
    \label{fig:tarot}
\end{figure}

We collected data in an indoor environment equipped with a Vicon motion capture system. This system provides precise six DoF position measurements of an aerial vehicle at $30$ Hz, as shown in Fig.~\ref{fig:gt-sample}. The capture area measures $5$ meters in width, $7.5$ meters in length, and $3$ meters in height. Fig.~\ref{fig:environment} shows the environment that includes various objects of different sizes placed randomly. The aerial vehicle is built on the Tarot $680$ Pro carbon fiber hexacopter frame and is equipped with an NVIDIA Jetson Nano, an Intel RealSense D435 depth camera, and Vicon 3D motion capture markers as shown in Fig.~\ref{fig:tarot}. The camera is mounted on the front of the vehicle. For simplicity and safety, the vehicle was handheld during data collection. While capturing images, the vehicle was moved around the environment at varying velocities. The Vicon system was carefully calibrated before the data was captured.
  
We captured $16$ sequences, resulting in a total of $28,523$ images at approximately $15$ frames per second. Additionally, for each sequence we generated $6$ more sequences with induced failures, leading to a total of $112$ sequences and $199,661$ images with $13,861.12$ seconds in total duration. RGB images were captured at a resolution of $1920$x$1080$, while depth images were captured at $640$x$480$. In the dataset, the depth and RGB images are aligned to match the same pixel positions. All scripts used for post-processing are publicly available\footref{footnote:repo}. In several sequences, one, two, or three people were asked to walk around the environment. Additionally, in some sequences, they were instructed to interact with each other or with objects in the environment. Finally, we captured sequences with the lights off or turned the lights off during recording. Table \ref{tab:sequences} summarizes the captured data.

\begin{table}
\centering
\caption{Dataset sequences details. Each sequence is augmented by creating 6 new versions with the proposed failure injection.}
\label{tab:sequences}
\begin{tabular}{|ccccc|}
\hline
Sequence & Duration (s) & Images & Dynamic & Low light \\ \hline
01       & 119.62       &  1732 & \checkmark & \xmark\\
02       & 166.12       &  2449 & \xmark     & \xmark  \\
03       & 139.57       &  2029 & \xmark    &  \xmark      \\
04       & 140.03       &  2053 & \checkmark & \xmark           \\
05       & 184.66       &  2689 & \checkmark & \xmark         \\
06       & 183.60       &  2677 & \checkmark & \xmark       \\
07       & 158.72       &  2309 & \checkmark & \xmark       \\
08       & 116.35       &  1705 & \xmark   &   \xmark      \\
09       & 87.13        &  1284 & \xmark   &   \xmark       \\
10       & 117.88       &  1743  & \xmark   &  \xmark        \\
11       & 97.14        &  1437 & \xmark   &  \xmark        \\
12       & 92.73        &  1372 & \xmark   &  \xmark        \\
13       & 94.47        &  1403 & \xmark   & \xmark          \\
14       & 78.99        &  831 & \xmark &\checkmark            \\
15       & 119.42       &  1608 & \xmark &\checkmark             \\
16       & 83.73        &  1202 & \xmark &\checkmark           \\ \hline
\end{tabular}
\end{table}

After data collection, we developed a script to inject various types of failures into the captured images, enhancing the dataset with six specific failure modes: camera underexposure (Fig. \ref{fig:underexposure}), camera overexposure (Fig. \ref{fig:overexposure}), broken lens (Fig. \ref{fig:breakage}), wet lens (Fig. \ref{fig:wet}), lens condensation (Fig. \ref{fig:condensation}), and lens dirt (Fig. \ref{fig:dirt}). The camera underexposure and overexposure failures were simulated using gamma power transformation, as described in \cite{lift-lfnet}. The other failure modes were generated by superimposing the original images with noise templates, following the method outlined in \cite{rgb-failures}. These failures have been shown to significantly impact the performance of visual odometry (VO) systems, as evaluated in \cite{camera-failures, modvo}. The scripts used to inject these failures into the sequences are also publicly available\footref{footnote:repo}. They allow users to easily introduce new types of failures into the dataset or apply failures to their own images.

\subsection{Data evaluation}
\label{subsection:evaluation}
To evaluate the data's integrity and quality, we performed experiments in all sequences of the dataset. For these experiments, we used two open-source algorithms: the RGB-D version of ORB-SLAM2\footnote{\label{ft-sg}\url{github.com/raulmur/ORB_SLAM2}} \cite{orb-slam2} and the TartanVO\footnote{\label{ft-sg}\url{github.com/castacks/tartanvo}} \cite{tartan-vo}.

\textbf{ORB-SLAM2} \cite{orb-slam2} is an open-source Visual SLAM library for Monocular, RGB-D and Stereo cameras. It is a feature-based VSLAM system that relies on Oriented FAST and rotated BRIEF (ORB) features and works with three threads: Tracking, Local Mapping, and Loop Closing, which allows the system to work with real-time performance \cite{orb-slam}. In addition, the mapping step adopts graph representations, which allows the system to perform local and global pose-graph optimization. It is one of the most popular and reliable RGB-D VSLAM systems.

\textbf{TartanVO} \cite{tartan-vo} is a Deep learning-based Visual Odometry algorithm. This model is trained using the TartanAir dataset \cite{tartanair}. In addition, they propose an up-to-scale loss function and incorporate the camera intrinsic parameters into the model. The results show that TartanVO, trained only on synthetic data without finetuning, can be applied to real-world datasets such as KITTI and Euroc MAV, providing significant advantages over geometry-based methods on challenging trajectories \cite{tartan-vo}.

We employed qualitative and quantitative approaches to assess the performance of the algorithms. The qualitative approach relied on visually comparing 3D plots of the predicted and ground-truth trajectories. The quantitative evaluation is based on the comparison between the predicted trajectories and the ground-truth trajectory. The metric used for this evaluation was the Absolute Trajectory Error (ATE), which evaluates the global accuracy of the predicted trajectories. The ATE can be computed as

\begin{equation}
    ATE = \frac{1}{N} \sum_{i=1}^{N} \left \|  \mathbf{\hat p}_i - \mathbf{p}_i  \right \|_{2},
    \label{eq:ate}
\end{equation}
where $N$ is the number of frames, $\mathbf{\hat{p}_i}$ is the estimated pose for frame $i$ and $\mathbf{p}_i$ is the ground-truth pose for the same frame. The trajectories must be aligned and on the same scale.

\section{RESULTS}
\label{sec:results}
In this section, we present the quantitative and qualitative results obtained with the experiments described in Section~\ref{subsection:evaluation}. All metrics and plots presented in this section were computed with a Python package to evaluate Odometry and SLAM called EVO \cite{evo}.

\subsection{Quantitative Results}

\begin{table*}
\centering
\caption{ATE (in meters) obtained with ORB-SLAM2 and TartanVO in QueensCAMP dataset. We do not present results in the sequences where ORB-SLAM2 cannot track at least 50\% of the trajectory. Highlighted values represent errors that were at least twice as big when a failure was added to a sequence when compared to the algorithm's performance with the original images.}
\label{tab:results}
\begin{tabular}{|ccccccccccccccccc|}
\hline
\multicolumn{17}{|c|}{ORB-SLAM2}                                                                                                                       \\ \hline
\multicolumn{1}{|c|}{Failure/Sequence} & 01   & 02   & 03   & 04   & 05   & 06   & 07   & 08   & 09   & 10   & 11   & 12   & 13   & 14   & 15   & 16   \\ \hline
\multicolumn{1}{|c|}{Original}         & 0.15 & 0.24 & 0.10 & 0.66 & 0.97 & 0.53 & 0.63 & -    & 0.15 & -    & 0.16 & 0.24 & 0.27 & 0.52 & 0.56 & 0.29 \\
\multicolumn{1}{|c|}{Underexposure}    & 0.16 & 0.23 & 0.12 & 0.73 & 0.89 & 0.62 & 0.57 & -    & 0.18 & -    & 0.16 & 0.27 & -    & -    & -    & 0.13 \\
\multicolumn{1}{|c|}{Overexposure}     & 0.20 & 0.18 & 0.09 & 0.77 & 1.17 & 0.63 & 0.82 & -    & 0.15 & -    & 0.18 & 0.13 & 0.21 & 0.36 & -    & 0.25 \\
\multicolumn{1}{|c|}{Breakage}         & \textbf{1.55} & - & \textbf{0.81} & -    & 1.15 & -    & -    & 1.26 & \textbf{1.69} & 2.32 & \textbf{1.16} & \textbf{2.06} & -    & \textbf{1.99} & \textbf{1.43} & -    \\
\multicolumn{1}{|c|}{Wet}              & \textbf{3.01} & \textbf{1.91} & \textbf{1.49} & \textbf{1.54} & 1.64 & \textbf{1.89} & \textbf{1.57} & 1.33 & \textbf{2.39} & 2.15 & \textbf{1.50} & \textbf{2.15} & \textbf{1.55} & \textbf{2.06} & \textbf{2.18} & \textbf{1.33} \\
\multicolumn{1}{|c|}{Condensation}     & -    & -    &\textbf{1.60} & -    & -    & -    & -    & 1.39 & \textbf{1.74} & 2.29 & -    & \textbf{2.00} & -    & \textbf{2.02} & \textbf{1.79} & \textbf{1.42} \\
\multicolumn{1}{|c|}{Dirt}             & 0.26 & -    & -    & 0.70 & -    & -    & -    & -    & \textbf{1.07} & -    & -    & -    & -    & -    & -    & -    \\ \hline
\multicolumn{17}{|c|}{TartanVO}                                                                                                                                                                                                                                    \\ \hline
\multicolumn{1}{|c|}{Failure/Sequence} & 01            & 02            & 03   & 04   & 05            & 06            & 07            & 08   & 09            & 10            & 11            & 12            & 13            & 14   & 15            & 16            \\ \hline
\multicolumn{1}{|c|}{Original}         & 0.87          & 0.38          & 0.90 & 1.13 & 0.80          & 0.63          & 0.63          & 0.70 & 0.45          & 0.57          & 0.55          & 0.75          & 0.61          & 0.94 & 0.97          & 0.63          \\
\multicolumn{1}{|c|}{Underexposure}    & 0.85          & 0.38          & 0.90 & 1.13 & 0.82          & 0.63          & 0.61          & 0.69 & 0.45          & 0.51          & 0.55          & 0.73          & 0.61          & 0.92 & 0.94          & 0.61          \\
\multicolumn{1}{|c|}{Overexposure}     & 0.91          & 0.38          & 0.89 & 1.11 & 0.81          & 0.63          & 0.65          & 0.69 & 0.45          & 0.56          & 0.54          & 0.75          & 0.60          & 0.99 & 0.97          & 0.66          \\
\multicolumn{1}{|c|}{Breakage}         & 1.37          & \textbf{0.87} & 1.28 & 1.07 & 1.08          & \textbf{1.61} & 1.08          & 0.96 & \textbf{1.36} & \textbf{1.43} & 0.89          & 1.27          & 1.10          & 1.75 & 1.81          & 1.04          \\
\multicolumn{1}{|c|}{Wet}              & \textbf{2.46} & \textbf{1.12} & 1.55 & 1.52 & \textbf{1.64} & \textbf{1.87} & \textbf{1.37} & 1.26 & \textbf{1.52} & \textbf{2.24} & \textbf{1.17} & \textbf{2.08} & 1.13          & 1.85 & \textbf{2.03} & \textbf{1.36} \\
\multicolumn{1}{|c|}{Condensation}     & \textbf{1.91} & \textbf{1.29} & 1.53 & 1.30 & 1.43          & \textbf{1.76} & \textbf{1.28} & 1.29 & \textbf{1.42} & \textbf{2.11} & \textbf{1.24} & \textbf{1.97} & 1.19          & 1.77 & \textbf{1.96} & \textbf{1.37} \\
\multicolumn{1}{|c|}{Dirt}             & 1.51          & 0.58          & 1.30 & 0.92 & 1.09          & 1.19          & 1.07          & 1.11 & \textbf{1.23} & \textbf{1.41} & 0.86          & 1.47          & \textbf{1.35} & 1.73 & 1.48          & 1.05          \\ \hline
\end{tabular}

\end{table*}
 Table \ref{tab:results} presents the ATE obtained using the proposed algorithms across all sequences in the dataset. For ORB-SLAM2, tracking can be lost when the number of extracted and matched keypoints is insufficient, with relocalization occurring only when the system recognizes a previously seen location \cite{orb-slam}. Therefore, we report results only for sequences where at least 50\% of the images were successfully tracked. In contrast, as a deep learning-based method, TartanVO does not extract keypoints from images and tracks all images regardless of the pose estimation quality.

Observing Table \ref{tab:results}, we found that ORB-SLAM2 fails to track at least 50\% of the trajectories in most scenarios involving lens dirt, and in half of the sequences with lens condensation. On the other hand, in sequences with wet lenses, although ORB-SLAM2 maintains trajectory tracking, the quality of the estimated trajectories is consistently lower compared to other scenarios. This degradation occurs because the algorithm continues to extract and match a sufficient number of keypoints, even when many of these are outliers, generally due to water droplets on the lens.

Additionally, we observed that ORB-SLAM2 failed to track the camera pose in sequences $08$ and $10$, which contain featureless areas viewed for extended periods. In these sequences, the algorithm did not lose track of the estimated trajectory with the breakage, wet, and condensation failures, however, the pose estimates present high errors. This is because, in the absence of sufficient distinctive features, the algorithm predominantly tracked and matched points generated by the lens failures.

Furthermore, we found that the injection of underexposure and overexposure failures did not significantly impact the performance of TartanVO, as the error remained consistent across all sequences, with or without these failures. For ORB-SLAM2, underexposure only negatively affected the algorithm's performance in sequences $13$ through $15$. The main reason for this degradation in sequences $14$ and $15$ is that the low light conditions were exacerbated by underexposure. Surprisingly, in some cases, the performance of both algorithms improved with underexposure and overexposure. This behavior was previously noted in \cite{lift-lfnet-slam, lift-slam}, the ill-exposure failures can remove regions of the images that contain outliers, potentially enhancing motion estimation accuracy in certain scenarios.

Moreover, whereas ORB-SLAM2 typically exhibited lower errors due to its more comprehensive optimization steps as a full VSLAM system, TartanVO showed greater consistency in a larger number of sequences, since its performance did not degrade by a factor of $2$ when analyzing all failures in $4$ sequences.

\subsection{Qualitative Results}

\begin{figure*}[thpb]
\centering
\subfloat[][ORB-SLAM2 trajectories in Sequence $01$.]{
\includegraphics[width=0.45\textwidth]{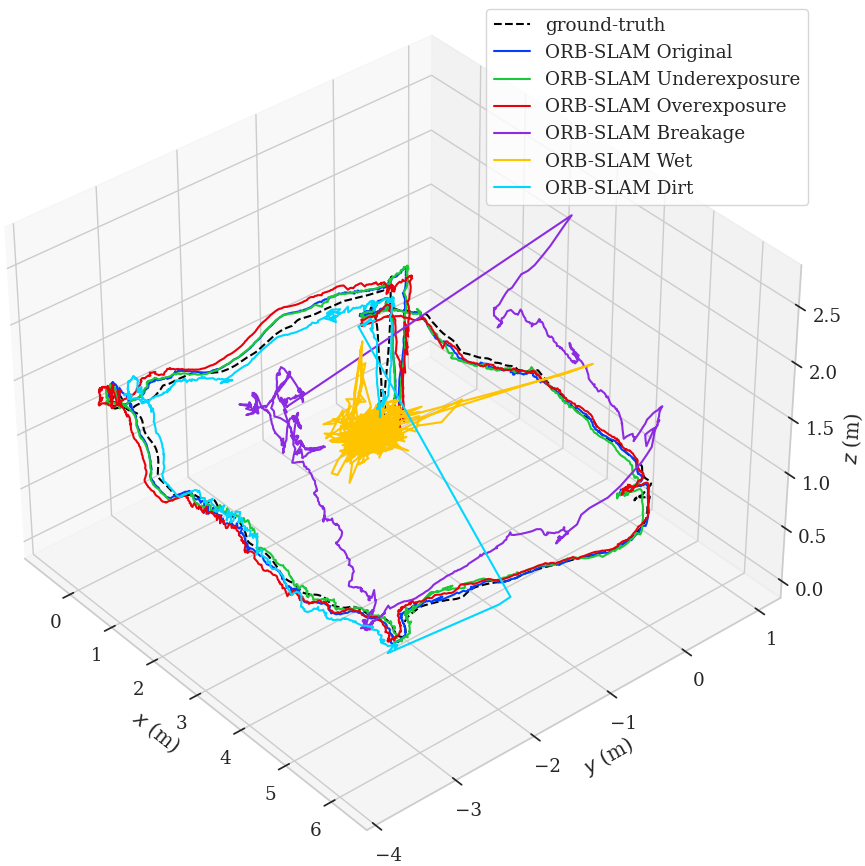}
\label{fig:results-orbslam-01}}
\qquad
\subfloat[][TartanVO trajectories in Sequence $01$.]{
\includegraphics[width=0.45\textwidth]{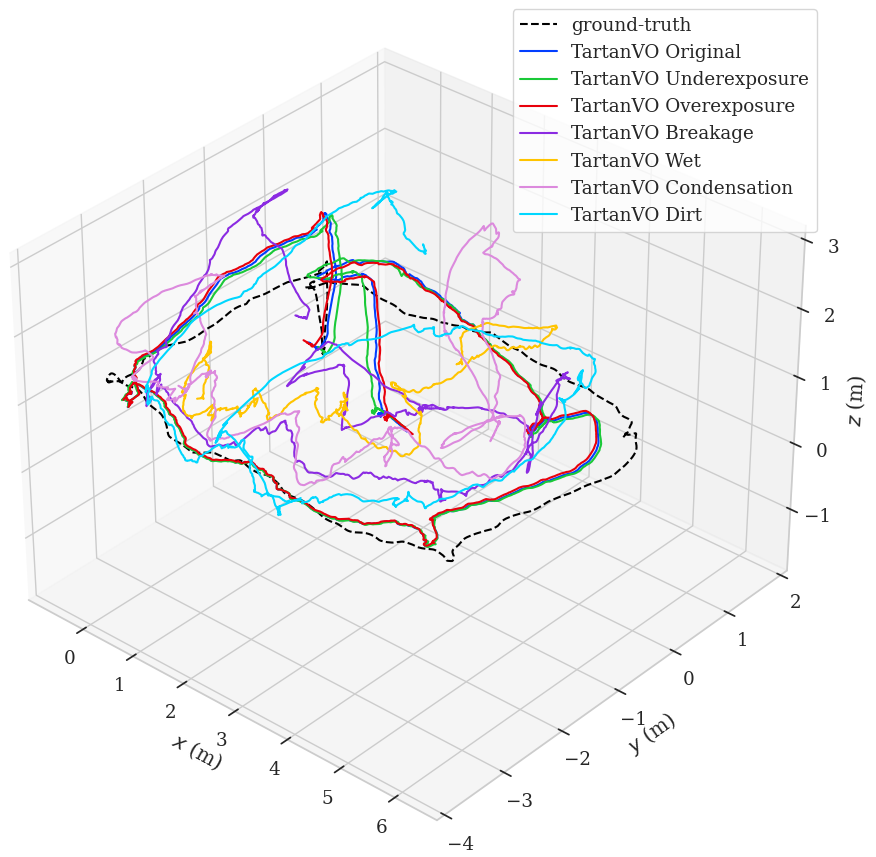}
\label{fig:results-tartanvo-01}}

\subfloat[][ORB-SLAM2 trajectories in Sequence $14$.]{
\includegraphics[width=0.45\textwidth]{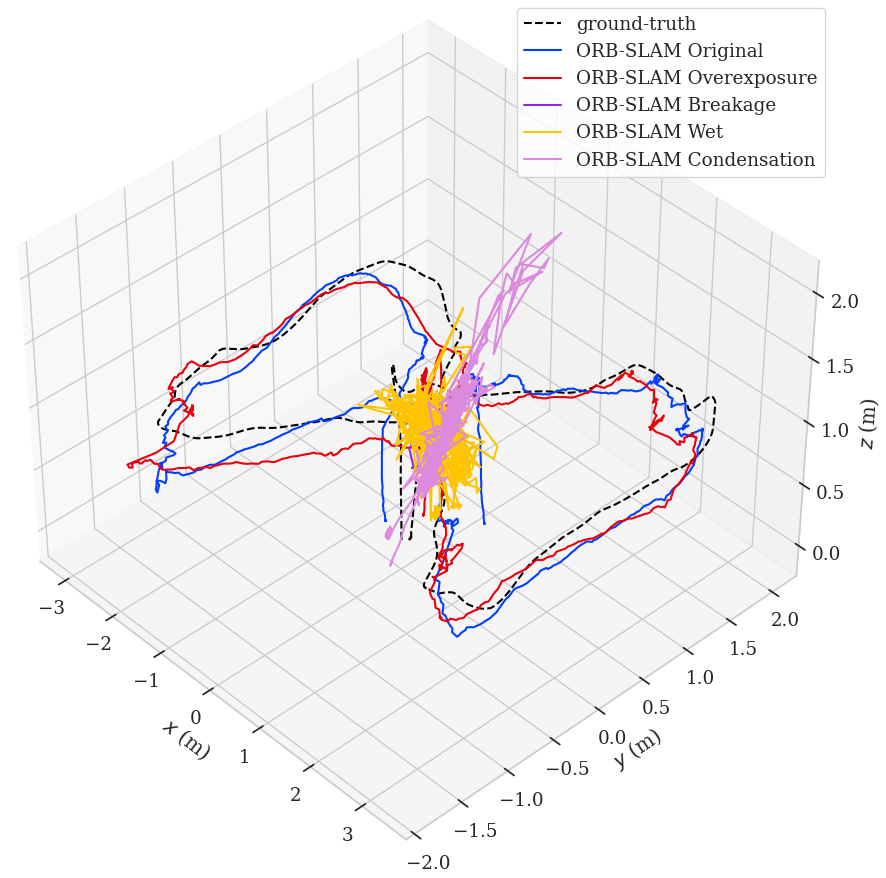}
\label{fig:results-orbslam-14}}
\qquad
\subfloat[][TartanVO trajectories in Sequence $14$.]{
\includegraphics[width=0.45\textwidth]{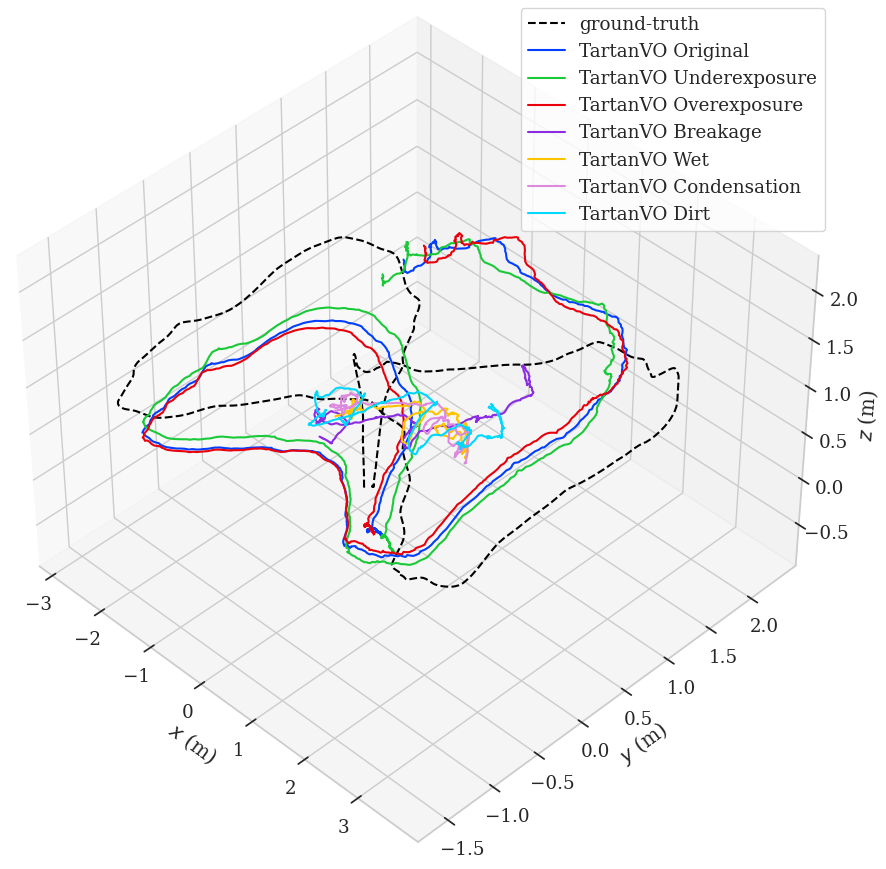}
\label{fig:results-tartanvo-14}}

\caption{Plots comparing the trajectories estimated by the algorithms with the ground-truth trajectory for sequences $01$ and $14$ of the dataset.}
\label{fig:results-plots}
\end{figure*}

We selected sequences $01$ and $14$ for the qualitative evaluation due to their distinct characteristics. Sequence $01$ features dynamic objects and favorable lighting conditions, as previously detailed in Table~\ref{tab:sequences}. In contrast, sequence $14$ is characterized by low light conditions and static objects. The plots comparing the estimated trajectories and ground-truth trajectory are shown in Fig.~\ref{fig:results-plots}.

In Fig.~\ref{fig:results-orbslam-01} the impact of breakage and wet lens failures on ORB-SLAM2 is evident. These failures significantly degrade the estimated trajectory, which aligns with the substantial errors observed in the quantitative results. Additionally, the pose estimation degradation caused by condensation is clearly visible in Fig. \ref{fig:results-orbslam-14}. Moreover, in sequence $01$ the dirt lens failure causes the algorithm to lose track of a significant portion of the trajectory. In sequence $14$, the lens breakage causes the algorithm to estimate minimal camera movement, rendering the trajectory almost static and invisible in the plot.

The trajectories estimated by TartanVO are consistent with the original estimation when applying underexposure and overexposure. Interestingly, the algorithm was robust to underexposure even in sequence $14$ (Fig. \ref{fig:results-tartanvo-14}) where the low light conditions are exacerbated by noise injection. However, the algorithm exhibits considerable degradation when the other failures are introduced. Therefore, although the algorithm appears more consistent based on the ATE, the plots reveal that the estimated trajectories can still be significantly compromised under failure conditions.

\section{CONCLUSIONS}
\label{sec:conclusion}

This paper introduces the QueensCAMP dataset, designed to evaluate the robustness of RGB-D VSLAM algorithms. The dataset comprises RGB images, depth images, and six DoF ground-truth trajectories. The captured data features real-world indoor scenes with dynamic objects, motion blur, varying illumination, and emulated camera failures, including lens dirt, condensation, underexposure, and overexposure. Additionally, we offer open-source scripts for simulating camera failures into any images.

We conducted experiments using ORB-SLAM2 \cite{orb-slam2} and TartanVO \cite{tartan-vo} on all dataset sequences to assess its integrity and quality. Both quantitative and qualitative metrics were employed to assess the accuracy of estimated trajectories relative to the ground-truth trajectory and to evaluate the impact of emulated failures compared to the scenarios with the original images. Our findings indicate that the addition of failures, particularly those that introduce significant artifacts (such as lens breakage, wet lenses, lens condensation, and lens dirt), significantly degrades the performance of the evaluated algorithms. Thus, our goal in providing this dataset and its tools is to offer the research community new ways to identify limitations and enhance the robustness of VO, VSLAM, and related algorithms.

\balance





\section*{ACKNOWLEDGMENT}
This work was supported by the Coordination for the Improvement of Higher Education Personnel (CAPES), Motorola, and Eldorado.
We would like to thank Daniel Franco, Dimitria Silveria, and Luiz Eugênio for assisting in the data capture.

\bibliography{references.bib}
\bibliographystyle{IEEEtran}

\end{document}